\documentclass[11pt]{article}
\usepackage{eacl2017}
\usepackage{times}
\usepackage{url}
\usepackage{latexsym}
\usepackage{amsmath}
\usepackage{multirow}
\usepackage{subfig}
\usepackage{microtype}
\usepackage{pgfplots}
\usepackage[subtle]{savetrees}

\eaclfinalcopy 
\def\eaclpaperid{174} 

\title{Character-Word LSTM Language Models}

\author{Lyan Verwimp\qquad Joris Pelemans\qquad Hugo Van hamme\qquad Patrick Wambacq \\
ESAT -- PSI, KU Leuven \\ 
Kasteelpark Arenberg 10, 3001 Heverlee, Belgium \\
   {\tt firstname.lastname@esat.kuleuven.be} \\}

\date{}

\begin{document}
\maketitle
\begin{abstract}
We present a Character-Word Long Short-Term Memory Language Model which both reduces the perplexity with respect to a baseline word-level language model and reduces the number of parameters of the model. Character information can reveal structural (dis)similarities between words and can even be used when a word is out-of-vocabulary, thus improving the modeling of infrequent and unknown words. By concatenating word and character embeddings, we achieve up to 2.77\% relative improvement on English compared to a baseline model with a similar amount of parameters and 4.57\% on Dutch. Moreover, we also outperform baseline word-level models with a larger number of parameters. 

\end{abstract}

\section{Introduction}

Language models (LMs) play a crucial role in many speech and language processing tasks, among others speech recognition, machine translation and optical character recognition. The current state of the art are recurrent neural network (RNN) based LMs \cite{Mik:10}, and more specifically long short-term memory models (LSTM)~\cite{Hoch} LMs~\cite{Sun:12} and their variants (e.g.\ gated recurrent units (GRU) \cite{Cho:14}). LSTMs and GRUs are usually very similar in performance, with GRU models often even outperforming LSTM models despite the fact that they have less parameters to train. However, Jozefowicz et al.~\shortcite{Joz:15} recently showed that for the task of language modeling LSTMs work better than GRUs, therefore we focus on LSTM-based LMs.

In this work, we address some of the drawbacks of NN based LMs (and many other types of LMs). A first drawback is the fact that the parameters for infrequent words are typically less accurate because the network requires a lot of training examples to optimize the parameters. The second and most important drawback addressed is the fact that the model does not make use of the internal structure of the words, given that they are encoded as one-hot vectors. 
For example, `felicity' (great happiness) is a relatively infrequent word (its frequency is much lower compared to the frequency of `happiness' according to Google Ngram Viewer~\cite{ngram}) and will probably be an out-of-vocabulary (OOV) word in many applications, but since there are many nouns also ending on `ity' (ability, complexity, creativity \ldots), knowledge of the surface form of the word will help in determining that `felicity' is a noun. Hence, subword information can play an important role in improving the representations for infrequent words and even OOV words. 

In our character-word (CW) LSTM LM, we concatenate character and word embeddings and feed the resulting character-word embedding to the LSTM. Hence, we provide the LSTM with information about the structure of the word. By concatenating the embeddings, the individual characters (as opposed to e.g.\ a bag-of-characters approach) are preserved and the order of the characters is implicitly modeled. Moreover, since we keep the total embedding size constant, the `word' embedding shrinks in size and is partly replaced by character embeddings (with a much smaller vocabulary and hence a much smaller embedding matrix), which decreases the number of parameters of the model.

We investigate the influence of the number of characters added, the size of the character embeddings, weight sharing for the characters and the size of the (hidden layer of the) model. Given that common or similar character sequences do not always occur at the beginning of words (e.g.\ `overfitting' -- `underfitting'), we also examine adding the characters in forward order, backward order or both orders. 

We test our CW LMs on both English and Dutch. Since Dutch has a richer morphology than English due to among others its productive compounding (see e.g.~\cite{Reveil}), we expect that it should benefit more from a LM augmented with formal/morphological information.

The contributions of this paper are the following:

\begin{enumerate}
\item We present a method to combine word and subword information in an LSTM LM: concatenating word and character embeddings. As far as we know, this method has not been investigated before. 

\item By decreasing the size of the word-level embedding (and hence the huge word embedding matrix), we effectively reduce the number of parameters in the model (see section~\ref{parameters}).
\item We find that the CW model both outperforms word-level LMs with the same number of hidden units (and hence a larger number of parameters) and word-level LMs with the same number of parameters. These findings are confirmed for English and Dutch, for a small model size and a large model size. The size of the character embeddings should be proportional to the total size of the embedding (the concatenation of characters should not exceed the size of the word-level embedding), and using characters in the backward order improves the perplexity even more (see sections~\ref{order},~\ref{english} and~\ref{dutch}). 

\item The LM improves the modeling of OOV words by exploiting their surface form (see section~\ref{oov}). 

\end{enumerate}

The remainder of this paper is structured as follows: first, we discuss related work (section~\ref{related}); then the CW LSTM LM is described (section~\ref{model}) and tested (section~\ref{exp}). Finally, we give an overview of the results and an outlook to future work (section~\ref{concl}).

\section{Related work}
\label{related}

Other work that investigates the use of character information in RNN LMs either completely replaces the word-level representation by a character-level one or combines word and character information. Much research has also been done on modeling other types of subword information (e.g.\ morphemes, syllables), but in this discussion, we limit ourselves to characters as subword information. 

Research on replacing the word embeddings entirely has been done for neural machine translation (NMT) by Ling et al.~\shortcite{Ling} and Costa-juss\`{a} and Fonollosa~\shortcite{Costa}, who replace word-level embeddings with character-level embeddings. Chung et al.~\shortcite{Chung} use a subword-level encoder and a character-level decoder for NMT. In dependency parsing, Ballesteros et al.~\shortcite{Ball} achieve improvements by generating character-level embeddings with a bidirectional LSTM. Xie et al.~\shortcite{Xie} work on natural language correction and also use an encoder-decoder, but operate for both the encoder and the decoder on the character level. 

Character-level word representations can also be generated with convolutional neural networks (CNNs), as Zhang et al.~\shortcite{Zhang} and Kim et al.~\shortcite{Kim} have proven for text classification and language modeling respectively. Kim et al.~\shortcite{Kim} achieve state-of-the-art results in language modeling for several languages by combining a character-level CNN with highway~\cite{highway} and LSTM layers. However, the major improvement is achieved by adding the highway layers: for a small model size, the purely character-level model without highway layers does not perform better than the word-level model (perplexity of 100.3 compared to 97.6), even though the character model has two hidden layers of 300 LSTM units each and is compared to a word model of two hidden layers of only 200 units (in order to keep the number of parameters similar). For a model of larger size, the character-level LM improves the word baseline (84.6 compared to 85.4), but the largest improvement is achieved by adding two highway layers (78.9). Finally, Jozefowicz et al.~\shortcite{Joz:16} also describe character embeddings generated by a CNN, but they test on the 1B Word Benchmark, a data set of an entirely different scale than the one we use.

Other authors combine the word and character information (as we do in this paper) rather than doing away completely with word inputs. Chen et al.~\shortcite{Chen} and Kang et al.~\shortcite{Kang} work on models combining words and Chinese characters to learn embeddings. Note however that Chinese characters more closely match subwords or words than phonemes. Bojanowski et al.~\shortcite{Boj} operate on the character level but use knowledge about the context words in two variants of character-level RNN LMs. Dos Santos and Zadrozny~\shortcite{Santos} join word and character representations in a deep neural network for part-of-speech tagging. Finally, Miyamoto and Cho~\shortcite{Miy} describe a LM that is related to our model, although their character-level embedding is generated by a bidirectional LSTM and we do not use a gate to determine how much of the word and how much of the character embedding is used. However, they only compare to a simple baseline model of 2 LSTM layers of each 200 hidden units without dropout, resulting in a higher baseline perplexity (as mentioned in section~\ref{english}, our CW model also achieves larger improvements than reported in this paper with respect to that baseline). 

We can conclude that in various NLP tasks, characters have recently been introduced in several different manners. However, the models investigated in related work are either not tested on a competitive baseline~\cite{Miy} or do not perform better than our models~\cite{Kim}. In this paper, we introduce a new and straightforward manner to incorporate characters in a LM that (as far as we know) has not been investigated before.  

\section{Character-Word LSTM LMs}
\label{model}

A word-level LSTM LM works as follows: a word encoded as a one-hot column vector $\textbf{w}_t$ (at time step $t$) is fed to the input layer and multiplied with the embedding matrix $\textbf{W}_{w}$, resulting in a word embedding $\textbf{e}_t$:

\begin{equation}
\label{eq:baseline}
\textbf{e}_t = \textbf{W}_{w} \times \textbf{w}_t
\end{equation}

The word embedding of the current word $\textbf{e}_t$ will be the input for a series of non-linear operations in the LSTM layer (we refer to~\cite{Zaremba} for more details about the equations of the LSTM cell). In the output layer, probabilities for the next word are calculated based on a softmax function.

\begin{figure}
\includegraphics[scale=0.95]{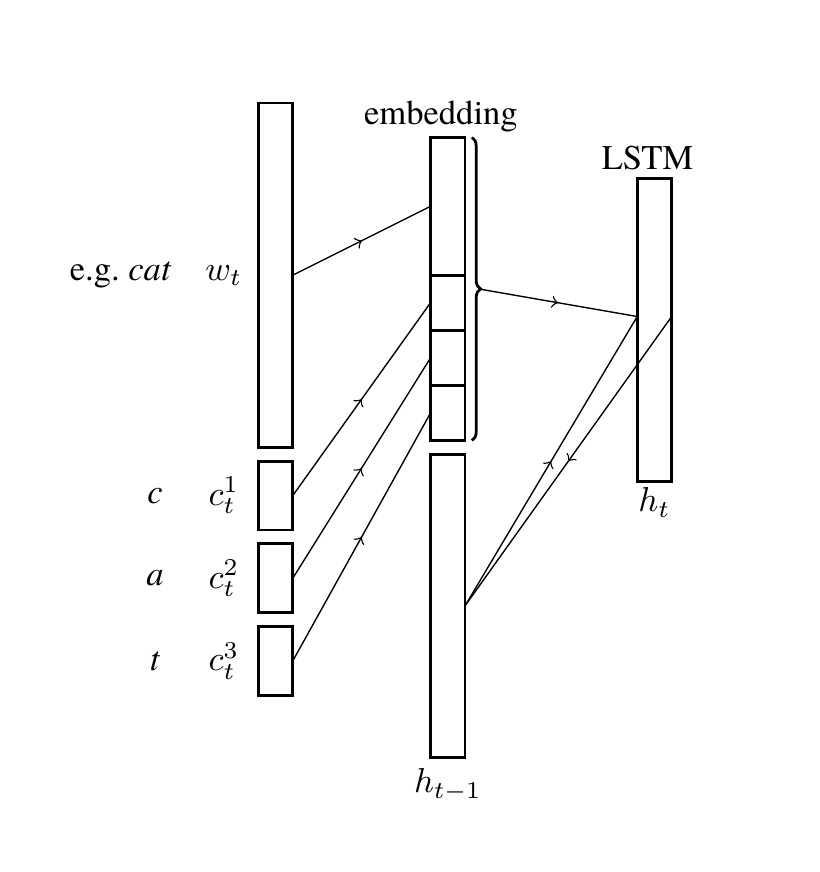}
\caption{Concatenating word and character embeddings in an LSTM LM.}
\label{CW}
\end{figure}

In our character-word LSTM LM, the only difference with the baseline LM is the computation of the `word' embedding, which is now the result of word and character input rather than word input only. We concatenate the word embedding with embeddings of the characters occurring in that word:

\begin{equation}
\label{eq:one-concat}
\begin{split}
\textbf{e}_t^\top = [(\textbf{W}_{w} \times \textbf{w}_t)^\top (\textbf{W}_{c}^1 \times \textbf{c}^1_{t})^\top \\
(\textbf{W}_{c}^2 \times \textbf{c}^2_t)^\top~\ldots~(\textbf{W}_{c}^n \times \textbf{c}^n_t)^\top] 
\end{split}
\end{equation}
where $\textbf{c}^1_t$ is the one-hot encoding of the first character added, $\textbf{W}_c^1$ its embedding matrix and $n$ the total number of characters added to the model.
The word $\textbf{w}_t$ and its characters $\textbf{c}^1_t, \textbf{c}^2_t \ldots \textbf{c}^n_t$  are each projected onto their own embeddings, and the concatenation of the embeddings is the input for the LSTM layer. By concatenating the embeddings, we implicitly preserve the order of the characters: the embedding for e.g.\ the first character of a word will always correspond to the same portion of the input vector for the LSTM  (see figure~\ref{CW}). We also experimented with adding word and character embeddings (a method which does not preserve the order of the characters), but that did not improve the perplexity of the LM. 

The number of characters added ($n$) is fixed. If a word is longer than $n$ characters, only the first (or last, depending on the order in which they are added) $n$ characters are added. If the word is shorter than $n$, it is padded with a special symbol. Because we can still model the surface form of OOV words with the help of their characters, this model reduces the number of errors made immediately after OOV words (see section~\ref{oov}).

\subsection{Order of the characters}
\label{order}

The characters can be added in the order in which they appear in the word (in the experiments this is called `forward order'), in the reversed order (`backward order') or both (`both orders'). In English and Dutch (and many other languages), suffixes can bear meaningful relations (such as plurality and verb conjugation) and compounds typically have word-final heads. Hence, putting more emphasis on the end of a word might help to better model those properties.

\subsection{Weight sharing}
\label{sharing}

Note that in equation~\ref{eq:one-concat} each position in the word is associated with different weights: the weights for the first character $\textbf{c}^1_t$, $\textbf{W}_{c}^1$, are different from the weights for the character in the second position, $\textbf{W}_{c}^2$. Given that the input `vocabulary' for characters is always the same, one could argue that the same set of weights $\textbf{W}_c$ could be used for all positions in the word:

\begin{equation}
\label{eq:concat-weights}
\begin{split}
\textbf{e}_t^\top = [(\textbf{W}_{w} \times \textbf{w}_t)^\top (\textbf{W}_{c} \times \textbf{c}^1_{t})^\top \\
(\textbf{W}_{c} \times \textbf{c}^2_t)^\top~\ldots~(\textbf{W}_{c} \times \textbf{c}^n_t)^\top] 
\end{split}
\end{equation}

However, one could also argue in favor of the opposite case (no shared weights between the characters): for example, an `s' at the end of a word often has a specific meaning, such as indicating a third person singular verb form of the present tense (in English), which it does not have at other positions in the word. Both models with and without weight sharing are tested (see section~\ref{weights-exp}). 

\subsection{Number of parameters}
\label{parameters}

Given that a portion of the total embedding is used for modeling the characters, the actual `word' embedding is smaller which reduces the number of parameters significantly. In a normal word-level LSTM LM, the number of parameters in the embedding matrix is

\begin{equation}
\label{size-word}
V \times E
\end{equation}
with $V$ the vocabulary size and $E = E_w$ the total embedding size/word embedding size. In our CW model however, the number of parameters is

\begin{equation}
\label{size-CW}
V \times (E - n \times E_c) + n \times (C \times E_c)
\end{equation}
with $n$ the number of characters, $E_c$ the size of the character embedding and $C$ the size of the character vocabulary. Since $V$ is by far the dominant factor, reducing the size of the purely word-level embedding vastly reduces the total number of parameters to train. If we share the character weights, that number becomes even smaller:

\begin{equation}
\label{size-CW-shared}
V \times (E - n \times E_c) + C \times E_c
\end{equation}

\section{Experiments}
\label{exp}

\subsection{Setup}
\label{setup}

All LMs were trained and tested with TensorFlow \cite{Tensorflow}. We test the performance of the CW architectures for a small model and a large model, with hyperparameters based on Zaremba et al.~\shortcite{Zaremba} and Kim et al.~\shortcite{Kim}). The small LSTM consists of 2 layers of 200 hidden units and the large LSTM has 2 layers of 650 hidden units. The total size of the embedding layer always equals the size of the hidden layer. During the first 4/6 (small/large model) epochs, the learning rate is 1, after which we apply an exponential decay:

\begin{equation}
\eta_{i} = \alpha~\eta_{i-1}
\label{lr}
\end{equation}
where $\eta_i$ is the learning rate at epoch $i$ and $\alpha$ the learning rate decay, which is set to 0.5 for the small LSTM and to 0.8 for the large LSTM. The smaller $\alpha$, the faster the learning rate decreases. The total number of epochs is fixed to 13/39 (small/large model). During training, 25\% of the neurons are dropped~\cite{dropout} for the small model and 50\% for the large model. The weights are randomly initialized to small values (between -0.1 and 0.1 for the small model and between -0.05 and 0.05 for the large model) based on a uniform distribution. We train on mini-batches of 20 with backpropagation through time, where the network is unrolled for 20 steps for the small LSTM and 35 for the large LSTM. The norm of the gradients is clipped at 5 for both models.

For English, we test on the publicly available Penn Treebank (PTB) data set, which contains 900k word tokens for training, 70k word tokens as validation set and 80k words as test set. This data set is small but widely used in related work (among others Zaremba et al.~\shortcite{Zaremba} and Kim et al.~\shortcite{Kim}), enabling the comparison between different models. We adopt the same pre-processing as used by previous work~\cite{Mik:10} to facilitate comparison, which implies that the dataset contains only lowercase characters (the size of the character vocabulary is 48). Unknown words are mapped to \textit{$\langle$unk$\rangle$}, but since we do not have the original text, we cannot use the characters of the unknown words for PTB.

The Dutch data set consists of components g, h, n and o of the Corpus of Spoken Dutch (CGN)~\cite{CGN}, containing recordings of meetings, debates, courses, lectures and read text. Approximately 80\% was chosen as training set (1.4M word tokens), 10\% as validation set (180k word tokens) and 10\% as test set (190k word tokens). The size of the Dutch data set is chosen to be similar to the size of the English data set. We also use the same vocabulary size as used for Penn Treebank (10k), since we want to compare the performance on different languages and exclude any effect of the vocabulary size. However, we do not convert all uppercase characters to lowercase (although the data is normalized such that sentence-initial words with a capital are converted to lowercase if necessary) because the fact that a character is uppercase is meaningful in itself. The character vocabulary size is 88 (Dutch also includes more accented characters due to French loan words, e.g. `caf\'{e}'). Hence, we do not only compare two different languages but also models with only lowercase characters and models with both upper- and lowercase characters. Moreover, since we have the original text at our disposal (as opposed to PTB), we can use the characters of the unknown words and still have a character-level representation. 

\subsection{Baseline models}
\label{baselines}

In our experiments, we compare the CW model with two word-level baselines: one with the same number of hidden units in the LSTM layers (thus containing more parameters) and one with approximately the same number of parameters as the CW model (like Kim et al.~\shortcite{Kim} do), because we are interested in both reducing the number of parameters and improving the performance. For the latter baseline, this implies that we change the number of hidden units from 200 to 175 for the small model and from 650 to 475 for the large, keeping the other hyperparameters the same. 

The number of parameters for those models is larger than for all CW models except when only 1 or 2 characters are added. The size difference between the CW models and the smaller word-level models becomes larger if more characters are added, if the size of the characters embeddings is larger and if the character weights are shared. The size of the embedding matrix for a word-level LSTM of size 475 is 10,000 $\times$ 475 = 475,000 ($V$ is 10k in all our experiments), whereas for a CW model with 10 character embeddings of size 25 it is of size 10,000 $\times$ (650 - 10 $\times$ 25) + 10 $\times$ (48 $\times$ 25) = 412,000 (the size of the character vocabulary for PTB is 48), following equation~\ref{size-CW}. If the character weights are shared, the size of the embedding matrix is only 401,200 (equation~\ref{size-CW-shared}).

The baseline perplexities for the smaller word-level models are shown in table~\ref{tab:baseline-small}. In the remainder of this paper, `w$x$' = means word embeddings of size $x$ for a word-level model and `c$x$' means character embeddings of size $x$ for CW models.

\begin{table}[h]
\centering
\begin{tabular}{l l l|c|c}
\hline
\multicolumn{3}{l}{}&\multicolumn{2}{c}{Perplexity} \\
\multicolumn{1}{l}{Corpus}&\multicolumn{2}{c}{Size}&\multicolumn{1}{c}{Validation}&Test \\
\hline
\multirow{4}{*}{PTB}&\multirow{2}{*}{small}&w200&100.7&96.86 \\
&&w175&102.62&98.82 \\
&\multirow{2}{*}{large}&w650&87.38&83.6 \\
&&w465&88.39&84.38  \\
\hline
\multirow{4}{*}{CGN}&\multirow{2}{*}{small}&w200&69.13&76 \\
&&w175&69.6&76.78 \\
&\multirow{2}{*}{large}&w650&63.36&70.69 \\
&&w475&63.88&70.88 \\
\hline
\end{tabular}
\caption{Perplexities for the baseline models. Baselines w200 and w650 have the same number of hidden units as the CW models and baselines w175 and w475 approximately have the same number of parameters as the CW models.}
\label{tab:baseline-small}
\end{table}

\subsection{English}
\label{english}

\begin{figure}[h]
\resizebox{7.8cm}{9cm}{\begin{tikzpicture}
\begin{axis}[
    xmin=1, xmax=10,
    ymin=99, ymax=106,
	xlabel={number of characters},
	ylabel={perplexity},
    legend style={at={(0.5,-0.2)},anchor=north}
]

\addplot[black,dashed,mark=circle*] table {small_baseline_175h.dat};
\addlegendentry{Small baseline w175}
\addplot[black,mark=circle*] table {small_baseline_dropout.dat};
\addlegendentry{Small baseline w200}
\addplot[black,mark=*] table {small_concat_5h_dropout.dat};
\addlegendentry{Small CW c5}
\addplot[black,mark=triangle*] table {small_concat_10h_dropout.dat};
\addlegendentry{Small CW c10}
\addplot[black,mark=square*] table {small_concat_15h_dropout.dat};
\addlegendentry{Small CW c15}
\end{axis}
\end{tikzpicture}}
\caption{Validation perplexity results on PTB, small model. Different sizes for the character embeddings are tested (`c5', `c10', `c15').}
\label{fig:concat_small}
\end{figure}
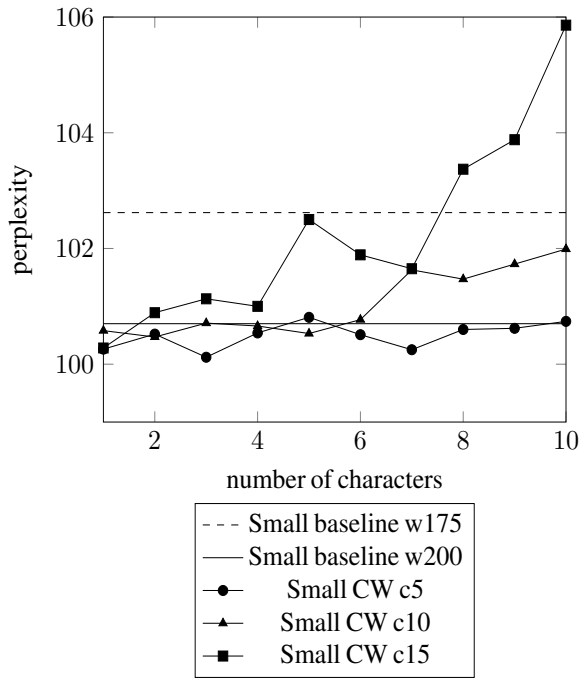

In figure~\ref{fig:concat_small}, the results for a small model trained on Penn Treebank are shown. Almost all CW models outperform the word-based baseline with the same number of parameters (2 LSTM layers of 175 units). Only the CW models in which the concatenated character embeddings take up the majority of the total embedding (more than 7 characters of embedding size 15) perform worse. With respect to the word-level LM with more parameters, only small improvements are obtained. The smaller the character embeddings, the better the performance of the CW model. For example, for a total embedding size of only 200, adding 8 character embeddings of size 15 results in an embedding consisting of 120 units `character embedding' and only 80 units `word embedding', which is not sufficient. The two best performing models add 3 and 7 character embeddings of size 5, giving a perplexity of 100.12 and 100.25 respectively, achieving a relative improvement of 2.44\%/2.31\% w.r.t.\ the w175 baseline and 0.58\%/0.45\% w.r.t.\ the w200 baseline. For those models, the `word embedding' consists of 185 and 165 units respectively. 

\begin{figure}[h]
\resizebox{7.8cm}{9cm}{\begin{tikzpicture}
\begin{axis}[
    xmin=1, xmax=10,
    ymin=85, ymax=89,
	xlabel={number of characters},
	ylabel={perplexity},
    legend style={at={(0.5,-0.2)},anchor=north}
]

\addplot[black,dashed,mark=circle*] table {large_baseline_475.dat};
\addlegendentry{Large baseline w475}
\addplot[black,mark=circle*] table {large_baseline.dat};
\addlegendentry{Large baseline w650}
\addplot[black,mark=*] table {large_concat_10h.dat};
\addlegendentry{Large CW c10}
\addplot[black,mark=triangle*] table {large_concat_25h.dat};
\addlegendentry{Large CW c25}
\addplot[black,mark=square*] table {large_concat_50h.dat};
\addlegendentry{Large CW c50}

\end{axis}
\end{tikzpicture}
\label{fig:concat-large-en-25-50}}
\caption{Validation perplexity results on PTB, large model. Different sizes for the character embeddings are tested (`c10', `c25', `c50').}
\label{fig:concat-large-en-size}
\end{figure}
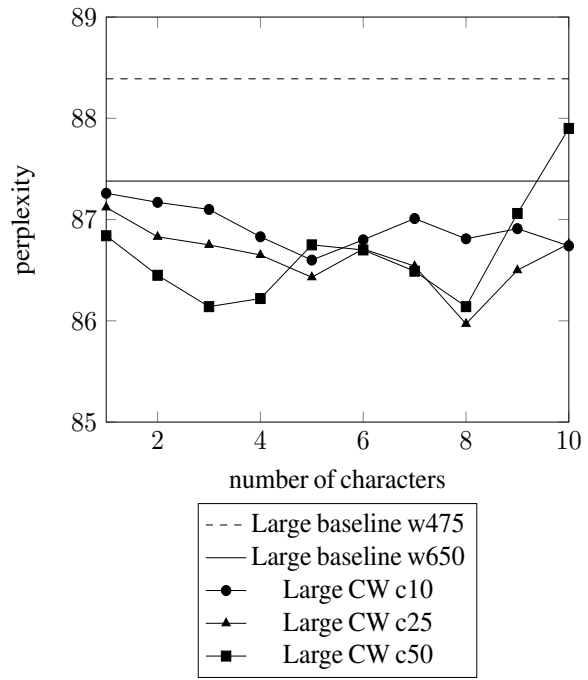

\begin{figure}[h]
\resizebox{7.7cm}{8.7cm}{\begin{tikzpicture}
\begin{axis}[
    xmin=1, xmax=15,
    ymin=85, ymax=89,
	xlabel={number of characters},
	ylabel={perplexity},
    legend style={at={(0.5,-0.2)},anchor=north}
]

\addplot[black,dashed,mark=circle*] table {large_baseline_475.dat};
\addlegendentry{Large baseline w475}
\addplot[black,mark=circle*] table {large_baseline.dat};
\addlegendentry{Large baseline w650}
\addplot[black,mark=*] table {large_concat_10h.dat};
\addlegendentry{Large CW c10 forward order}
\addplot[black,mark=triangle*] table {large_concat_10h_inverted.dat};
\addlegendentry{Large CW c10 backward order}
\addplot[black,mark=square*] table {large_concat_10h_both.dat};
\addlegendentry{Large CW c10 both orders}

\end{axis}
\end{tikzpicture}
\label{fig:concat-large-en-10}}
\caption{Validation perplexity results on PTB, large model. Several options for the order in which the characters are added are investigated.}
\label{fig:concat-large-en-order}
\end{figure}
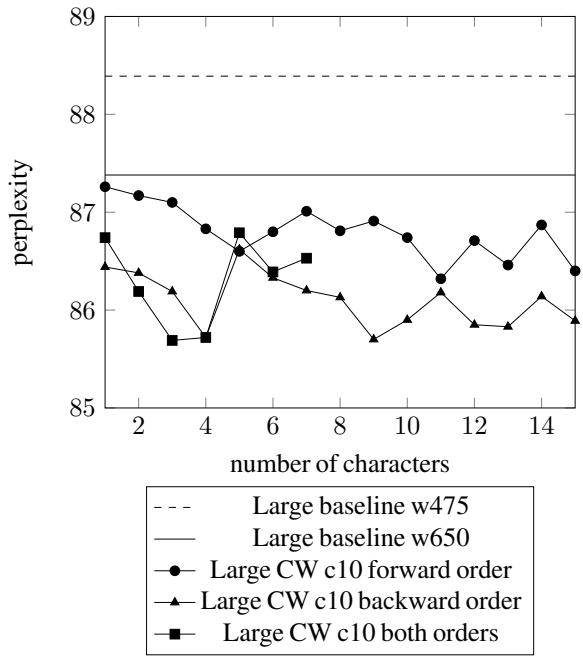

We test the performance of the CW architecture on a large model too. In figure~\ref{fig:concat-large-en-size}, the results for different embedding sizes are shown. Just like we saw for the small model, the size of the character embeddings should not be too large: for embeddings of size 50 (`c50'), the performance drops when a larger number of characters is added.
The best result is obtained by adding 8 characters with embeddings of size 25 (`c25'): a perplexity of 85.97 (2.74\%/1.61\% relative improvement with respect to the w475/w650 baseline). For embeddings of size 10, adding more than 10 characters gives additional improvements (see figure~\ref{fig:concat-large-en-order}).

We also verify whether the order in which the characters are added is important (figure~\ref{fig:concat-large-en-order}). The best result is achieved by adding the first 3 and the last 3 characters to the model (`both orders'), giving a perplexity of 85.69, 3.05\%/1.87\% relative improvement with respect to the w475/w650 baseline. However, adding more characters in both orders causes a decrease in performance. When only adding the characters in the forward order or the backward order, adding the characters in backward order seems to perform slightly better overall (best result: adding 9 characters in the backward order gives a perplexity of 85.7 or 3.04\%/1.92\% improvement with respect to the w475/w650 baseline). 

We can conclude that the size of the character embeddings should be proportional to the total embedding size: the word-level embedding should be larger than the concatenation of the character-level embeddings. Adding characters in the backward order is slightly better than adding them in the forward order, and the largest improvement is made for the large LSTM LM. The test perplexities for some of the best performing models (table~\ref{tab:ptb-test}) confirm these findings.

\begin{table}[h]
\centering
\begin{tabular}{l|c}
\hline
\textbf{Small model}&\textbf{Perplexity} \\
\hline
Baseline w175/w200 &98.82/96.86 \\
\hline
\cite{Kim}&100.3 \\
\cite{Miy}&109.05 \\
c5 with $n = 3$&\textbf{96.21} \\
c5 with $n = 7$&96.35 \\
\hline
\textbf{Large model}&\textbf{Perplexity} \\
\hline
Baseline w475/w650&84.38/83.6 \\
\hline
\cite{Kim}&84.6 \\
c25 with $n = 8$&82.69 \\
c10 with $n = 9(b)$&82.68 \\
c10 with $n = 3+3(b)$&\textbf{82.04} \\
\hline
\end{tabular}
\caption{Test perplexity results for  the best models on PTB. Baseline perplexities are for sizes w175/w200 for a small model and w475/w650 for a large model. $n$ = number of characters added, $(b)$ means backward order. Comparison with other character-level LMs~\cite{Kim} (we only compare to models without highway layers) and character-word models~\cite{Miy} (they do not use dropout and only report results for a small model).}

\label{tab:ptb-test}
\end{table}

If we compare the test perplexities with two related models that incorporate characters, we see that our models perform better. Kim et al.~\shortcite{Kim} generate character-level embeddings with a convolutional neural network and also report results for both a small and a large model. Their small character-level model has more hidden units than ours (300 compared to 200), but it does not improve with respect to the word-level baseline (since we do not use highway layers, we only compare with the results for models without highway layers). Their large model slightly improves their own baseline perplexity (85.4) by 0.94\%.\ Compare with our results: 2.64\% perplexity reduction for the best small LSTM (c5 with $n = 3$) and 2.77\% for the best large LSTM (c10 with $n = 3+3(b)$). Miyamoto and Cho~\shortcite{Miy} only report results for a small model that is trained without dropout, resulting in a baseline perplexity of 115.65. If we train our small model without dropout we get a comparable baseline perplexity (116.33) and a character-word perplexity of 110.54 (compare to 109.05 reported by Miyamoto and Cho~\shortcite{Miy}). It remains to be seen whether their model performs equally well compared to better baselines. Moreover, their hybrid character-word model is more complex than ours because it uses a bidirectional LSTM to generate the character-level embedding (instead of a lookup table) and a gate to determine the mixing weights between the character- and word-level embeddings.

\subsection{Dutch}
\label{dutch}

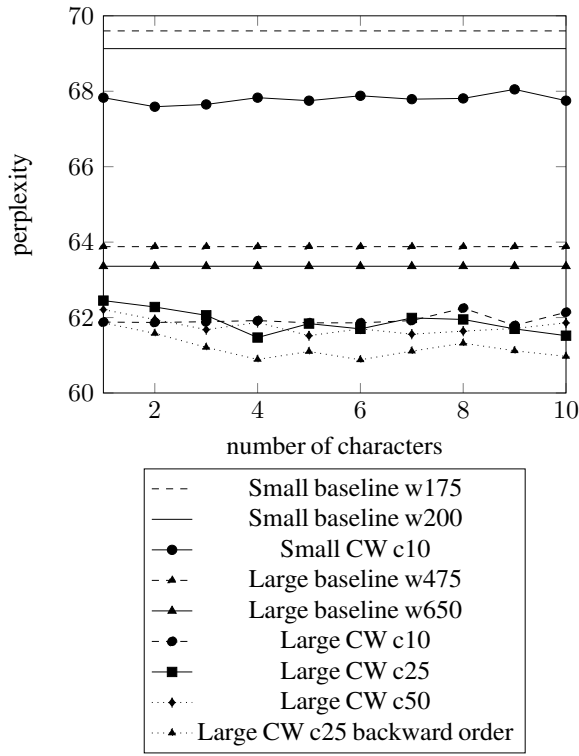
\begin{figure}[h]
\centering
\resizebox{7.8cm}{10cm}{\begin{tikzpicture}
\begin{axis}[
    xmin=1, xmax=10,
    ymin=60, ymax=70,
	xlabel={number of characters},
	ylabel={perplexity},
    legend style={at={(0.5,-0.2)},anchor=north}
]
\addplot[black,dashed,mark=circle*] table {cgn_small_baseline_175h.dat};
\addlegendentry{Small baseline w175}
\addplot[black,mark=circle*] table {cgn_small_baseline.dat};
\addlegendentry{Small baseline w200}
\addplot[black,mark=*] table {cgn_small_append.dat};
\addlegendentry{Small CW c10}


\addplot[black,dashed,mark=triangle*] table {cgn_large_baseline_475h.dat};
\addlegendentry{Large baseline w475}
\addplot[black,mark=triangle*] table {cgn_large_baseline.dat};
\addlegendentry{Large baseline w650}
\addplot[black,dashed,mark=*] table {cgn_large_append.dat};
\addlegendentry{Large CW c10}
\addplot[black,mark=square*] table {cgn_large_append_25h.dat};
\addlegendentry{Large CW c25}
\addplot[black,dotted,mark=diamond*] table {cgn_large_append_50h.dat};
\addlegendentry{Large CW c50}
\addplot[black,dotted,mark=triangle*] table {cgn_large_append_25h-inv.dat};
\addlegendentry{Large CW c25 backward order}

\end{axis}
\end{tikzpicture}}
\caption{Validation perplexity results on CGN. Several options for the size and order of the character embeddings are investigated.}
\label{fig:concat-nl}
\end{figure}

As we explained in the introduction, we expect that using information about the internal structure of the word will help more for languages with a richer morphology. Although Dutch is still an analytic language  (most grammatical relations are marked with separate words or word order rather than morphemes), it has a richer morphology than English because compounding is a productive and widely used process and because it has more lexical variation due to inflection (e.g.\ verb conjugation, adjective inflection). 
The results for the LSTM LMs trained on Dutch seem to confirm this hypothesis (see figure~\ref{fig:concat-nl}).

The CW model outperforms the baseline word-level LM both for the small model and the large model. The best result for the small model is obtained by adding 2 or 3 characters, giving a perplexity of 67.59 or 67.65 which equals a relative improvement of 2.89\%/2.23\% (w175/w200) and 2.80\%/2.14\% (w175/w200) respectively. 

For the large model, we test several embedding sizes and orders for the characters. The best model is the one to which 6 characters in backward order are added, with a perplexity of 60.88 or 4.70\%/3.91\% (w475/w650) relative improvement. Just like for PTB, an embedding size of 25 proves to be the best compromise: if the characters are added in the normal order, 4 characters with embeddings of size 25 is the best model (perplexity 61.47 or 3.77\%/2.98\% (w475/w650) relative improvement).

These results are confirmed for the test set (table~\ref{tab:cgn-test}). The best small model has a perplexity of 75.04 which is 2.27\% compared to the baseline and the best large model has a perplexity of 67.64, a relative improvement of 4.57\%. The larger improvement for Dutch might be due to the fact that it has a richer morphology and/or the fact that we can use the surface form of the OOV words for the Dutch data set (see sections~\ref{setup} and~\ref{oov}).

\begin{table}[h]
\centering
\begin{tabular}{l|c}
\hline
\textbf{Small model}&\textbf{Perplexity} \\
\hline
Baseline w175/w200&76.78/76 \\
\hline
c10 with $n = 2$&75.23 \\
c10 with $n = 3$&\textbf{75.04} \\
\hline
\textbf{Large model}&\textbf{Perplexity} \\
\hline
Baseline w475/w650&70.88/70.69 \\
\hline
c25 with $n = 4$&68.79 \\
c25 with $n = 6(b)$&\textbf{67.64} \\
\hline
\end{tabular}
\caption{Test perplexity results for the best models on CGN. Baseline perplexities are for sizes w175/w200 for a small model and w475/w650 for a large model. $n$ = number of characters added, $(b)$ means backward order.}
\label{tab:cgn-test}
\end{table}

\subsection{Random CW models}

In order to investigate whether the improvements of the CW models are not caused by the fact that the characters add some sort of noise to the input, we experiment with adding real noise -- random `character' information -- rather than the real characters. Both the number of characters (the length of the random `word') and the `characters' themselves are generated based on a uniform distribution. In table~\ref{tab:random}, the relative change in perplexity, averaged over models to which 1 to 10 characters are added, with respect to the baseline word-level LM and the CW model with real characters is shown. 

\begin{table}[h]
\centering
\begin{tabular}{l|l|c|c}
\hline
\multicolumn{2}{l}{}&\multicolumn{2}{c}{Relative change in} \\
\multicolumn{2}{l}{}&\multicolumn{2}{c}{valid perplexity w.r.t.} \\
\multicolumn{2}{l}{}&\multicolumn{1}{c}{\textbf{Baseline}}&\textbf{Char-Word} \\
\hline
\multicolumn{1}{l}{\multirow{2}{*}{PTB}}&small c5&0.34 (0.30)&0.54 (0.46)  \\
\multicolumn{1}{l}{}&large c15&0.00 (0.29)&0.53 (0.49)\\
\hline
\multicolumn{1}{l}{\multirow{2}{*}{CGN}}&small c10&- 0.18 (0.53)&1.79 (0.47) \\
\multicolumn{1}{l}{}&large c10&- 0.15 (0.26)&1.52 (1.24) \\
\hline
\end{tabular}
\caption{Relative change in validation perplexity for models to which \textbf{random} information is added, w.r.t.\ word-level and CW models. The improvements are averaged over the results for adding 1 to 10 characters/random information, the numbers between brackets are standard deviations. Negative numbers indicate a decrease in perplexity.}
\label{tab:random}
\end{table}

For English, adding random information had a negative impact on the performance with respect to both the baseline and the CW model. For Dutch on the other hand, adding some random noise to the word-level model gave small improvements. However, the random models perform much worse than the CW models.\ We can conclude that the characters provide meaningful information to the LM.

\subsection{Sharing weights}
\label{weights-exp}

We repeat certain experiments with the CW models, but with embedding matrices that are shared across all character positions (see section~\ref{sharing}). Note that sharing the weights does not imply that the position information is lost, because for example the first portion of the character-level embedding always corresponds to the character on the first position. Sharing the weights ensures that a character is always mapped onto the same embedding, regardless of the position of that character in the word, e.g.\ both occurrences of `i' in `felicity' are represented by the same embedding. This effectively reduces the number of parameters.

We compare the performance of the CW models with weight sharing with the baseline word-level LM and the CW model without weight sharing.\ In table~\ref{tab:sharing}, the relative change with respect to those LMs is listed. 

CW models with weight sharing generally improve with respect to a word-level baseline, except for the small English LM. For Dutch, the improvements are more pronounced. The difference with the CW model without weight sharing is small (right column), although \textit{not} sharing the weights works slightly better, which suggests that characters can convey different meanings depending on the position in which they occur. Again, the results are more clear-cut for Dutch than for English. 

\begin{table}[h]
\centering
\begin{tabular}{l|l|c|c}
\hline
\multicolumn{2}{l}{}&\multicolumn{2}{c}{Relative change in} \\
\multicolumn{2}{l}{}&\multicolumn{2}{c}{valid perplexity w.r.t.} \\
\multicolumn{2}{l}{}&\multicolumn{1}{c}{\textbf{Baseline}}&\textbf{Char-Word} \\
\hline
\multicolumn{1}{l}{\multirow{2}{*}{PTB}}&small c10&0.53 (0.88)&0.19 (0.67) \\
\multicolumn{1}{l}{}&large c10&- 0.54 (0.37)&- 0.02 (0.22) \\
\hline
\multicolumn{1}{l}{\multirow{2}{*}{CGN}}&small c10&- 1.70 (0.34)&0.24 (0.30) \\
\multicolumn{1}{l}{}&large c10&- 2.10 (0.32)&0.15 (0.50) \\
\hline
\end{tabular}
\caption{Relative change in validation perplexity for CW models with \textbf{weight sharing} for the characters, w.r.t.\ baseline and CW models without weight sharing. The improvements are averaged over the results for adding 1 until 10 characters, the numbers between brackets are standard deviations. Negative numbers indicate a decrease in perplexity.}
\label{tab:sharing}
\end{table}

\subsection{Dealing with out-of-vocabulary words}
\label{oov}

As we mentioned in the introduction, we expect that by providing information about the surface form of OOV words (namely, their characters), the number of errors induced by those words should decrease. We conduct the following experiment to check whether this is indeed the case: for the CGN test set, we keep track of the probabilities of each word during testing. If an OOV word is encountered, we check the probability of the target word given by a word-level LM and a CW LM. The word-level model is a large model of size 475 and the CW model is a large model in which 6 characters embeddings of size 25 in the backward order are used (the best performing CW model in our experiments). 

We observe that in 17,483 of the cases, the CW model assigns a higher probability to the target word following an OOV word, whereas the opposite happens only in 10,724 cases. This is an indication that using the character information indeed helps in better modeling the OOV words. 

\section{Conclusion and future work}
\label{concl}

We investigated a character-word LSTM language model, which combines character and word information by concatenating the respective embeddings. This both reduces the size of the LSTM and improves the perplexity with respect to a baseline word-level LM. The model was tested on English and Dutch, for different model sizes, several embedding sizes for the characters, different orders in which the characters are added and for weight sharing of the characters. We can conclude that for almost all setups, the CW LM outperforms the word-level model, whereas it has fewer parameters than the word-level model with the same number of LSTM units. If we compare with a word-level LM with approximately the same number of parameters, the improvement is larger. 

One might argue that using a CNN or an RNN to generate character-level embeddings is a more general approach to incorporate characters in a LM, but this model is simple, easier to train and smaller. Moreover, related models using a CNN-based character embedding~\cite{Kim} do not perform better.

For both English and Dutch, we see that the size of the character embedding is important and should be proportional to the total embedding size: the total size of the concatenated character embeddings should not be larger than the word-level embedding. 
Not using the characters in the order in which they appear in the word, but in the reversed order (and hence putting more emphasis on the end of the word), performs slightly better, although adding only a few characters both from the beginning and the end of the word achieves good performance too. 

Using random inputs instead of the characters performed worse than using the characters themselves, thus refuting the hypothesis that the characters simply introduce noise. Sharing the weights/embedding matrices for the characters reduces the size of the model even more, but causes a small increase in perplexity with respect to a model without weight sharing. Finally, we observe that the CW models are better able to deal with OOV words than word-level LMs.

In future work, we will test other architectures to incorporate character information in a word-level LSTM LM, such as combining a character-level LSTM with a word-level LSTM.
Another representation that might be useful uses character co-occurrence vectors (by analogy with the acoustic co-occurrences used by Van hamme~\shortcite{HAC,HAC2}) rather than one-hot character vectors, because co-occurrences intrinsically give information about the order of the characters. Other models could be more inspired by human language processing: according to the theory of \textit{blocking}, we humans have both a mental lexicon of frequent words and a morphological module that is used to process infrequent/ unknown words or to create new words (see e.g.~\cite{blocking}). This could correspond to a word-level LM for frequent words and a subword-level LM for infrequent words.

\section*{Acknowledgments}

This research is funded by the Flemish government agency IWT (project 130041, SCATE).

\bibliography{eacl2017}
\bibliographystyle{eacl2017}

\end{document}